\documentclass[conference]{IEEEtran}
\IEEEoverridecommandlockouts
\usepackage{cite}
\usepackage{amsmath,amssymb,amsfonts}
\usepackage{algorithmic}
\usepackage{graphicx}
\usepackage{textcomp}
\usepackage{xcolor}
\usepackage{multirow}
\usepackage{makecell}
\def\BibTeX{{\rm B\kern-.05em{\sc i\kern-.025em b}\kern-.08em
    T\kern-.1667em\lower.7ex\hbox{E}\kern-.125emX}}
\begin{document}

\title{Uncertainty-Aware Cross-Modal Transfer Network for Sketch-Based 3D Shape Retrieval\thanks{* Corresponding author.}
}

% \author{Yiyang Cai, Jiaming Lu, Jiewen Wang, Shuang Liang}

% \address{School of Software Engineering, Tongji University, China}

\author{\IEEEauthorblockN{Yiyang Cai, Jiaming Lu, Jiewen Wang, Shuang Liang\IEEEauthorrefmark{1}}
\IEEEauthorblockA{School of Software Engineering, Tongji University, China \\
\{2131486, 2231522, wjwlaservne, shuangliang\}@tongji.edu.cn}
}

\maketitle

\begin{abstract}
In recent years, sketch-based 3D shape retrieval has attracted growing attention. While many previous studies have focused on cross-modal matching between hand-drawn sketches and 3D shapes, the critical issue of how to handle low-quality and noisy samples in sketch data has been largely neglected. This paper presents an uncertainty-aware cross-modal transfer network (UACTN) that addresses this issue. UACTN decouples the representation learning of sketches and 3D shapes into two separate tasks: classification-based sketch uncertainty learning and 3D shape feature transfer. We first introduce an end-to-end classification-based approach that simultaneously learns sketch features and uncertainty, allowing uncertainty to prevent overfitting noisy sketches by assigning different levels of importance to clean and noisy sketches. Then, 3D shape features are mapped into the pre-learned sketch embedding space for feature alignment. Extensive experiments and ablation studies on two benchmarks demonstrate the superiority of our proposed method compared to state-of-the-art methods.
\end{abstract}


\section{Introduction}
With the rapid growth in the number of 3D shapes in recent years, 3D shape retrieval has been studied extensively. Compared with other query forms, sketch-based methods are more intuitive and convenient for users to retrieve 3D shapes. Hence, sketch-based 3D shape retrieval (SBSR) has gained growing attention in the fields of computer vision ~\cite{2013SHREC,2014SHREC, Wang_2015_CVPR}.

Most of the previous work~\cite{Wang_2015_CVPR, Dai2018-ce, Chen2018-up, Lei2019-ki,dai2020cross,zhao2022jfln} has focused on the most obvious challenge of cross-modal matching between sketches and 3D shapes. These works have designed various network architectures and loss functions to map sketch and 3D shape features into a common embedding space. Another research focus has been on 3D shape representation, with many efforts~\cite{Xie2017-vb,Lei2019-ki,2019hashing,Xu2020} attempting to obtain better 3D shape representations to reduce the modality gap between sketches and 3D shapes. 

\begin{figure}[tbp]
\centering
\includegraphics[width=0.48\textwidth]{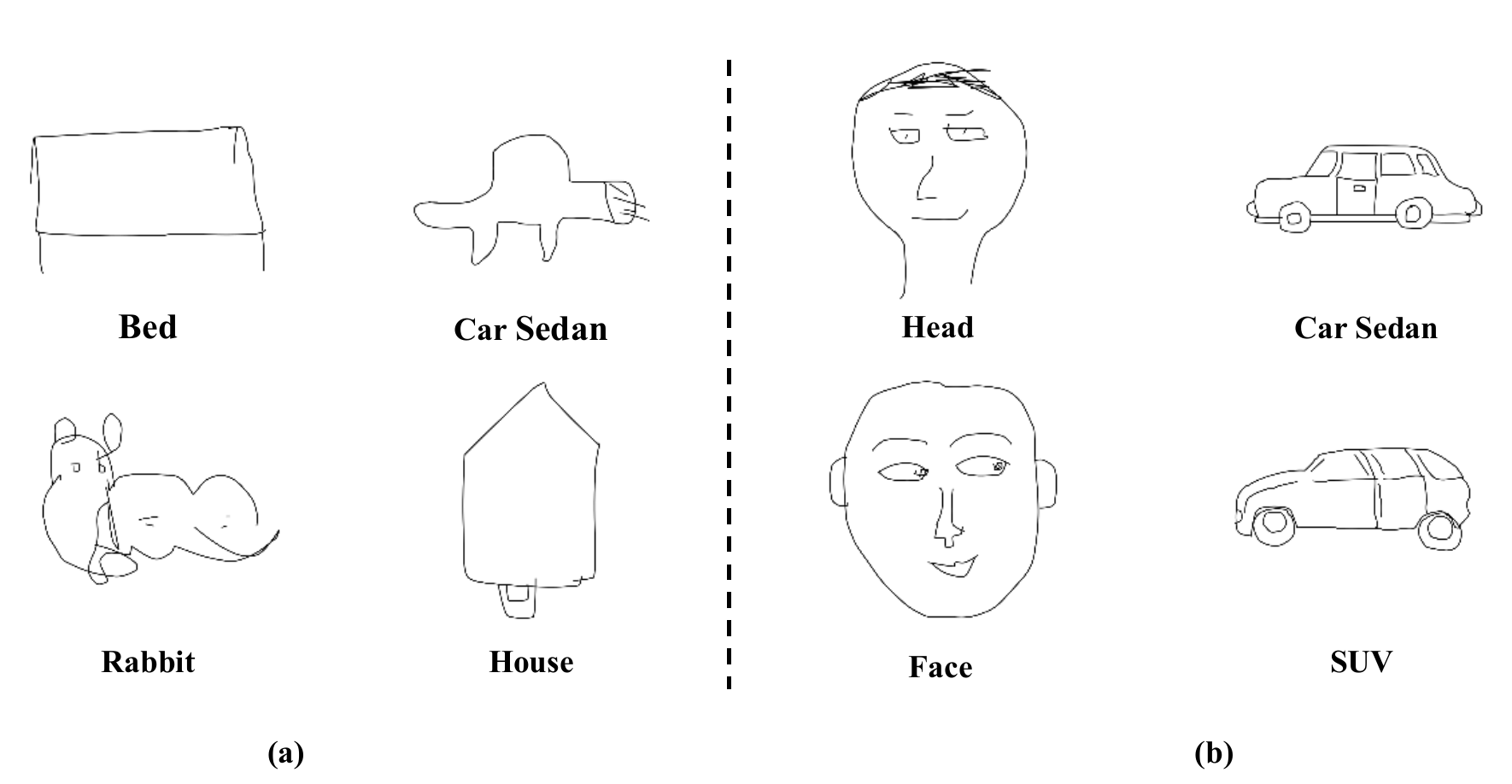}
\caption{Examples of high-quality clean sketches and low-quality noisy sketches from SHREC 2013.}
\label{fig1}
\end{figure}

There is still a lack of research on sketch representation learning in SBSR, with most work treating sketches as natural images. However, as a visual language that is highly abstract and lacks detail, sketches are more challenging to represent than natural images and often contain low-quality, noisy samples. Sketches can vary in their level of abstraction and detail, and some sketches (e.g., Fig.~\ref{fig1}) are so abstract that they are unrecognizable even to humans. These unrecognizable sketches are detrimental to model training, as the model will attempt to overfit noisy samples and learn irrelevant information. SUL~\cite{Liang2021-kg} is the only work focused on this issue and proposes a regression-based sketch uncertainty estimation approach to prevent the model from overfitting noisy sketch samples with high uncertainty. However, they separate sketch representation learning and uncertainty learning into two steps and use uncertainty to fine-tune the sketch branches on the trained model. Hence, only a few layers are tuned, limiting the retrieval performance improvement.

\begin{figure*}[htbp]
\centering
\includegraphics[width=0.98\textwidth]{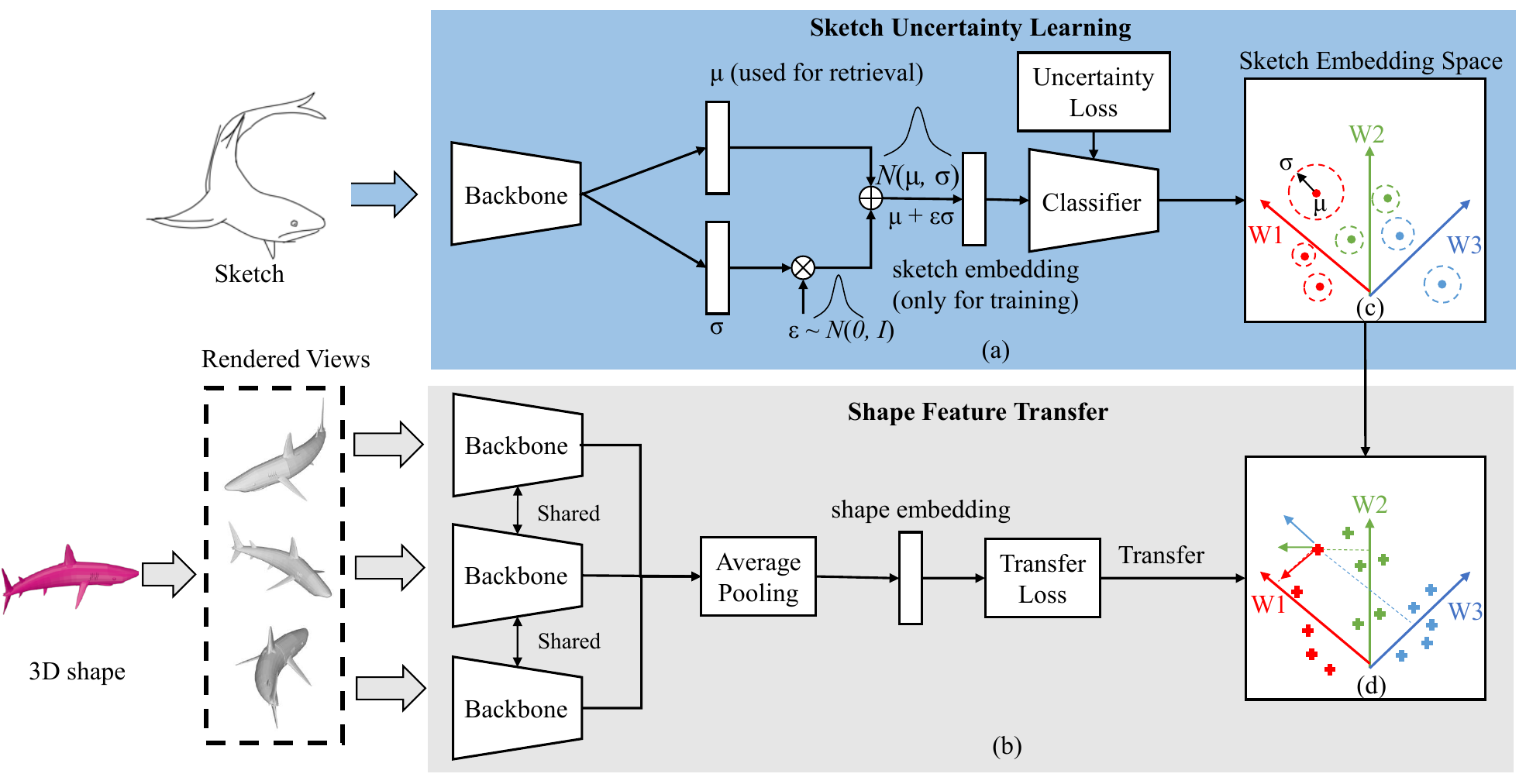}
\caption{The overview of our UACTN framework.}
\label{fig2}
\end{figure*}

\textbf{Our contributions.} In this paper, we propose an end-to-end uncertainty learning approach for sketch representation models that uses uncertainty to train the model from scratch, which addresses the limitations in SUL~\cite{Liang2021-kg}. Specifically, our improvements over SUL are reflected in the following aspects:

\begin{itemize}
    \item We propose a classification-based sketch uncertainty learning method, CBUL. Instead of employing uncertainty as a weighting parameter in the loss, we represents the sketch embedding and uncertainty as a probabilistic embedding so as to employ the classification loss to learn both the sketch representation and the uncertainty from scratch. By doing so, all network parameters and class center distribution in the embedding space are optimized by the uncertainty, which offers a more effective way of sketch uncertainty learning than SUL.
    \item We propose a novel framework called the Uncertainty-aware Cross-modal Transfer Network (UACTN). UACTN is a two-stage cross-modal matching method, which leverages transfer learning to integrate the proposed CBUL with cross-modal matching. It decouples the representation learning of sketches and 3D shapes into two separate steps to facilitate the use of the proposed uncertainty learning method in SBSR. Furthermore, the framework is able to achieve competitive results even without the proposed sketch uncertainty learning due to its ability to learn better class-discriminative embeddings by decoupling sketch and model representation learning.
    \item We conduct extensive experiments and ablation studies on widely used benchmarks, demonstrating the superiority of our proposed method compared to state-of-the-arts.
\end{itemize}

% The proposed uncertainty learning method is classification-based, introducing uncertainty into the training of the classification model to generate noise-robust discriminative embeddings of sketches. Based on this uncertainty learning method, we propose a novel framework called the Uncertainty-aware Cross-modal Transfer Network (UACTN). UACTN decouples the representation learning of sketches and 3D shapes into two separate steps to facilitate the use of the proposed uncertainty learning method in SBSR. Furthermore, the framework is able to achieve competitive results even without the proposed sketch uncertainty learning due to its ability to learn better class-discriminative embeddings by decoupling sketch and model representation learning. Finally, we conduct extensive experiments and ablation studies on widely used benchmarks, demonstrating the superiority of our proposed method compared to state-of-the-art methods.

\section{Related Work}

\noindent\textbf{Sketch-based 3D Shape Retrieval.} Sketch-based 3D shape retrieval (SBSR) is a challenging task that has been studied for many years. Early works proposed various methods based on handcrafted features~\cite{2013SHREC,2014SHREC,furuya2013ranking}, but deep learning methods~\cite{Wang_2015_CVPR, Xu2020, Qi2018SemanticEF,dai2020cross, Liang2021-kg} have become increasingly popular due to their superior performance. Wang et al.~\cite{Wang_2015_CVPR} are among the first to apply siamese networks and the contrastive loss for cross-modal matching between sketches and 3D shapes. Xu et al.~\cite{Xu2020} propose a view selection algorithm to find the most representative viewpoints. Qi et al.~\cite{Qi2018SemanticEF} propose cross-modal matching in a joint semantic embedding space, using classification-based learning for SBSR for the first time. Lei et al.~\cite{Lei2019-ki} propose a method with an improved center loss which combines classification-based loss and metric-based loss. Dai et al.~\cite{dai2020cross} propose a two-stage method for learning a common embedding space via knowledge distillation, which inspired us to decouple the representation learning of sketches and 3D shapes for better representation learning and scalability.

\noindent\textbf{Data Uncertainty Learning.} In deep learning, we can represent a data sample $x_i$ as an embedding $z_{i} = f(x_{i}) + n(x_{i})$. Here, $f(x_{i})$ denotes the ideal discriminative embedding, which mostly represents the semantic information of $x_{i}$, and $n(x_{i})$ denotes the uncertainty information of $x_{i}$. Data uncertainty learning aims to estimate the uncertainty information $n(x_{i})$ in $x_{i}$. One approach to achieving data uncertainty estimation is to represent the data sample as a Gaussian distribution rather than a fixed vector in the embedding space. The mean $\mu$ of the distribution denotes the most representative embedding $f(x_{i})$, and the variance $\sigma$ models the uncertainty information $n(x_{i})$ in the data sample $x_{i}$.

In recent years, data uncertainty is attracting more attention in various fields, including face recognition~\cite{chang2020data}, person ReID~\cite{Yu_2019_ICCV}, etc. Liang et al.~\cite{Liang2021-kg} first propose a regression-based uncertainty learning method to reduce the impact of noisy sketch data in training in the field of SBSR. This paper proposes an end-to-end sketch uncertainty learning approach to exploit uncertainty comprehensively.

\section{METHODOLOGY}

\subsection{Network Architecture}

The overall architecture of the proposed uncertainty-aware cross-modal transfer network (UACTN) for SBSR is illustrated in Fig.~\ref{fig2}. We decouple the task of cross-modal matching between sketches and 3D shapes into two separate learning tasks: (1) sketch data uncertainty learning, which aims to obtain a noise-robust sketch feature extraction model by introducing sketch uncertainty information into the training of a classification model; and (2) 3D shape feature transfer, where 3D shape features are mapped into the sketch embedding space under the guidance of sketch class centers. Finally, a cross-domain discriminative embedding space (i.e., sketches and 3D shapes belonging to the same class are close, while those of different classes are apart) is learned. The two tasks are discussed in detail in the following subsections.

In the retrieval phase, the features of query sketches and gallery 3D shapes are extracted using the models obtained in the two learning steps. The cosine similarity of the query sketch features and gallery 3D shape features is then calculated and ranked to obtain the retrieval results.

\subsection{Sketch Uncertainty Learning}
\label{sec3.2}

\textbf{Probabilistic Embedding.} To introduce the uncertainty information into sketch representation learning, we represent the sketch feature as a probabilistic embedding. Specifically, the embedding $z_{i}$ of a sketch sample $x_{i}$ is defined as a Gaussian distribution $\mathcal{N}(\mu_{i},\sigma^{2}_{i}I)$. Here the mean $\mu_{i}$ and the variance $\sigma^{2}_{i}$ are determined by $x_{i}$. Both $\mu_{i}$ and $\sigma_{i}$ are high dimensional vectors, where $\mu_{i}$ denotes the ideal class-discriminative embedding and $\sigma_{i}$ denotes the uncertainty of $\mu_{i}$. To obtain $\mu_{i}$ and $\sigma_{i}$, we first use a CNN backbone to extract the image feature of $x_{i}$ and then feed the feature into two separate fully connected networks to predict $\mu_{i}$ and $\sigma_{i}$. As illustrated in Fig.~\ref{fig2} (c), The sketch representation can be regarded as an embedding randomly sampled from $\mathcal{N}(\mu_{i},\sigma^{2}_{i}I)$. However, adding sampling operations to the model can prevent backpropagation. To address this issue, we use the reparameterization method in VAE~\cite{kingma2013auto}, which is illustrated in Fig.~\ref{fig2} (a). Instead of sampling directly from $\mathcal{N}(\mu_{i},\sigma^{2}_{i}I)$, we first sample a random vector $\epsilon$ from $\mathcal{N}(0,I)$, and then generate $z_{i}$ as the equivalent probabilistic representation:

\begin{equation}
\label{eq1}
z_{i}=\mu_{i}+\epsilon\cdot\sigma_{i},\quad\epsilon\sim\mathcal{N}(0,I) 
\end{equation}

\noindent With this method, we decouple sampling from the backpropagation workflow, thus enabling backpropagation. It is noted that the probabilistic embedding $z_{i}$ is exclusively used for training, while the discriminative embedding $\mu_{i}$ is used for similarity computation in retrieval.

\noindent\textbf{Loss Function.} Now $z_{i}$ is the probabilistic embedding of the sketch $x_{i}$ during training. $z_{i}$ is then fed to a classifier and optimized by the Large Margin Cosine Loss (LMCL)~\cite{wang2018cosface}:

\begin{equation}
\label{eq2}
\mathcal{L}_{lmc} = -\frac{1}{N}\sum_{i=1}^{N}\log\
\frac{e^{s(\overline{w}_{yi}\cdot\overline{z}_{i}-m_{s})}}{e^{s(\overline{w}_{yi}\cdot\overline{z}_{i}-m_{s})} + \sum_{j \neq yi}^{c}e^{s(\overline{w}_{j} \cdot \overline{z}_{i})}}
\end{equation}

\noindent Here, $N$ is the number of training samples, $C$ is the number of classes, and $\overline{w}_{yi}=\frac{w_{yi}}{||w_{yi}||}$, $\overline{z}_{i}=\frac{z_{i}}{||z_{i}||}$ are the normalization vectors of $w_{yi}$ and $z_{i}$, respectively. $w_{j}$ denotes the weight vector of the $j^{th}$ class from the final fully connected layer of the classifier, which can be regarded as the class center vector of the $j^th$ class. ${y_{i}}$ denotes the corresponding ground-truth label of the sample $z_{i}$. The parameter $s$ is used to control the convergence speed of the loss, and $m_{s}$ is the cosine margin that separates the decision boundaries of different classes. In our experiments, we set $s = 30$ and $m_{s} = 0.5$. The mechanism of $\mathcal{L}_{lmc}$ is to reduce the angle between ${z}_{i}$ and ${w}_{yi}$ and increase the angle with the other ${w}_{j}$, making it well-suited for cosine similarity based retrieval methods.

In order to suppress the uncertainty in $z_{i}$, the model will tend to predict a small and constant value for $\sigma$ for all sketch samples. However, this results in the probabilistic embedding $z_{i}$ being degraded to the fixed embedding $\mu_{i}$. To address this issue, a regularization loss is introduced to provide balance. The idea is to ensure that $\mathcal{N}(\mu_{i},\sigma^{2}{i}I)$ is close to a normal Gaussian distribution $\mathcal{N}(0,I)$. This is achieved by introducing Kullback-Leibler divergence to constrain $\mathcal{N}(\mu{i},\sigma^{2}_{i}I)$.

\begin{eqnarray}
\mathcal{L}_{kl} &=& D_{KL}(\mathcal{N}(\mu_{i},\sigma^{2}_{i}I)||\mathcal{N}(0,I)) \nonumber\\
~ &=& -\frac{1}{2}(1+\log\sigma^{2}-\mu^{2}-\sigma^{2})
\end{eqnarray}

\noindent $\mathcal{L}_{kl}$ is a monotonically decreasing function with respect to $\sigma$ under the condition that $\sigma^{(l)}_{i} \in (0,1)$ ($l$ denotes the $l^{th}$ dimension of $\sigma_{i}$). The final loss is $\mathcal{L}_{uncer}=\mathcal{L}_{lmc} + \lambda \mathcal{L}_{kl}$. Here $\lambda$ is a hyper-parameter set to 0.005 in our experiments. $\mathcal{L}_{uncer}$ converges each dimension of $\sigma_{i}$ to the range $(0,1)$.

\noindent\textbf{Mechanism Explanation.} Obviously, there are two questions regarding $\mathcal{L}_{uncer}$. (1) \textit{Why the model learns large variances for noisy samples?} It is noted that decreasing $sigma_{i}$ will decrease $\mathcal{L}_{lmc}$ and increase $\mathcal{L}_{kl}$. It is also noted that noisy sketch samples could make it difficult to decrease their $\mathcal{L}_{lmc}$ due to their semantic ambiguity. Now it is clear that decreasing $\sigma_{i}$ of noisy samples will increase $\mathcal{L}_{kl}$ but still lead to large $\mathcal{L}_{lmc}$ while decreasing those of clean samples will decrease $\mathcal{L}_{lmc}$ easily. In this case, decreasing $\sigma_{i}$ for clean samples leads to smaller $\mathcal{L}_{uncer}$. Hence, the model learns relatively larger $\sigma_{i}$ to noisy samples. (2) \textit{Why samples with larger variance could contribute less to model training?} The reason is that larger $\sigma_{i}$ will affect more severely the $\mu_{i}$, making the $z_{i}$ farther away from the original $\mu_{i}$ in embedding space. Hence, $z_{i}$ with larger $\sigma_{i}$ is more random and represents less information, preventing the model from overfitting noisy samples.

\subsection{3D Shape Feature Transfer}

\textbf{3D Shape Representation.} To map 3D shapes into the sketch embedding space, we first need to represent 3D shapes as features with the same dimensions as sketch features. Specifically, following MVCNN~\cite{Su2015-hp}, we adopt a multi-view-based approach to represent 3D shapes. We render a 3D shape into 12 views from different perspectives by evenly placing 12 virtual cameras around the 3D shape. A CNN backbone extracts features from the rendered views of the shape. An average pooling layer is used to fuse the view features, and the fused feature is fed into a fully connected network to match its dimension to the sketch feature.

\noindent\textbf{Transfer Loss.} The pre-learned sketch class centers are utilized to guide the learning of shape features, which map 3D shape features to the previously learned sketch embedding space. The transfer loss is formulated as follows:

\begin{equation}
\mathcal{L}_{t} = -\frac{1}{N}\sum_{i=1}^{N}\log\
\frac{e^{s(\overline{w}_{yi}\cdot\overline{f}_{i}-m_{v})}}{e^{s(\overline{w}_{yi}\cdot\overline{f}_{i}-m_{v})} + \sum_{j \neq yi}^{c}e^{s(\overline{w}_{j} \cdot \overline{f}_{i})}}
\end{equation}

\noindent Here, $\overline{f}_{i}$ and $\overline{w}_{yi}$ denote the normalization vectors of the shape embedding $f_{i}$ and the corresponding class center ${w}_{y}$ pre-learned in sketch uncertainty learning. $N$, $C$, $s$ and $m_{v}$ have the same meaning as in $\mathcal{L}_{lmc}$. Additionally, $s$ is set to 15, and $m_{v}$ is set to 0.8. $L_{t}$ has the same form as $L_{lmc}$, with the exception that the weight vectors $[w_{1},w_{2},...,w_{c}]$ are derived from pre-trained sketch weights and are fixed during training. As illustrated in Fig.~\ref{fig2} (d), the mechanism of $L_{t}$ is to cluster the shape features $f_{i}$ toward the class center $w_{yi}$ of the sketch in the same class, while also pushing the features away from the class centers $w_{j}$ of different classes.

\begin{table}
  \caption{The performance (\%) on SHREC 2013. For each metric, the best result under the same backbone is in bold.}
  \label{tab1}
  \resizebox{0.48\textwidth}{!}{
  \begin{tabular}{llllllll}
    \hline
    Method&Backbone&NN&FT&ST&E&DCG&mAP\\
    \hline
    DPSML\cite{Lei2019-ki}&ResNet50&81.9&83.4&87.5&41.5&89.2&85.7\\
    \hline
    CGN\cite{dai2020cross}&ResNet50&83.2&85.3&\textbf{90.2}&41.9&90.1&87.0\\
    \hline
    JFLN\cite{zhao2022jfln}&ResNet50&84.0&85.8&89.9&42.3&89.7&86.6\\
    \hline
    \multirow{2}{*}{DSSH\cite{2019hashing} } &ResNet50&79.9&81.4&86.0&40.4&87.3&83.1\\
    &I-R-v2&83.1&84.4&88.6&41.1&89.3&85.8\\
    \hline
    \multirow{2}{*}{HEAR\cite{chen2020learning}} &ResNet50&82.1&83.7&87.8&40.9&88.8&85.4\\
    &I-R-v2&84.2&85.6&88.8&41.3&90.0&86.9\\
    \hline
    \multirow{2}{*}{SUL\cite{Liang2021-kg}} 
    &ResNet50&82.4&84.3&89.3&41.7&89.6&86.2\\
    &I-R-v2&84.5&85.8&90.0&42.0&90.3&87.1\\
    \hline
    \multirow{2}{*}{\makecell[l]{UACTN\\(Ours)}}&ResNet50&\textbf{84.3}&\textbf{85.8}&89.9&\textbf{42.3}&\textbf{90.2}&\textbf{87.3}\\
    &I-R-v2&\textbf{85.4}&\textbf{87.1}&\textbf{90.9}&\textbf{42.8}&\textbf{91.3}&\textbf{88.6}\\
    \hline
\end{tabular}}
\end{table}

\begin{figure}[tbp]
\centering
\includegraphics[width=0.48\textwidth]{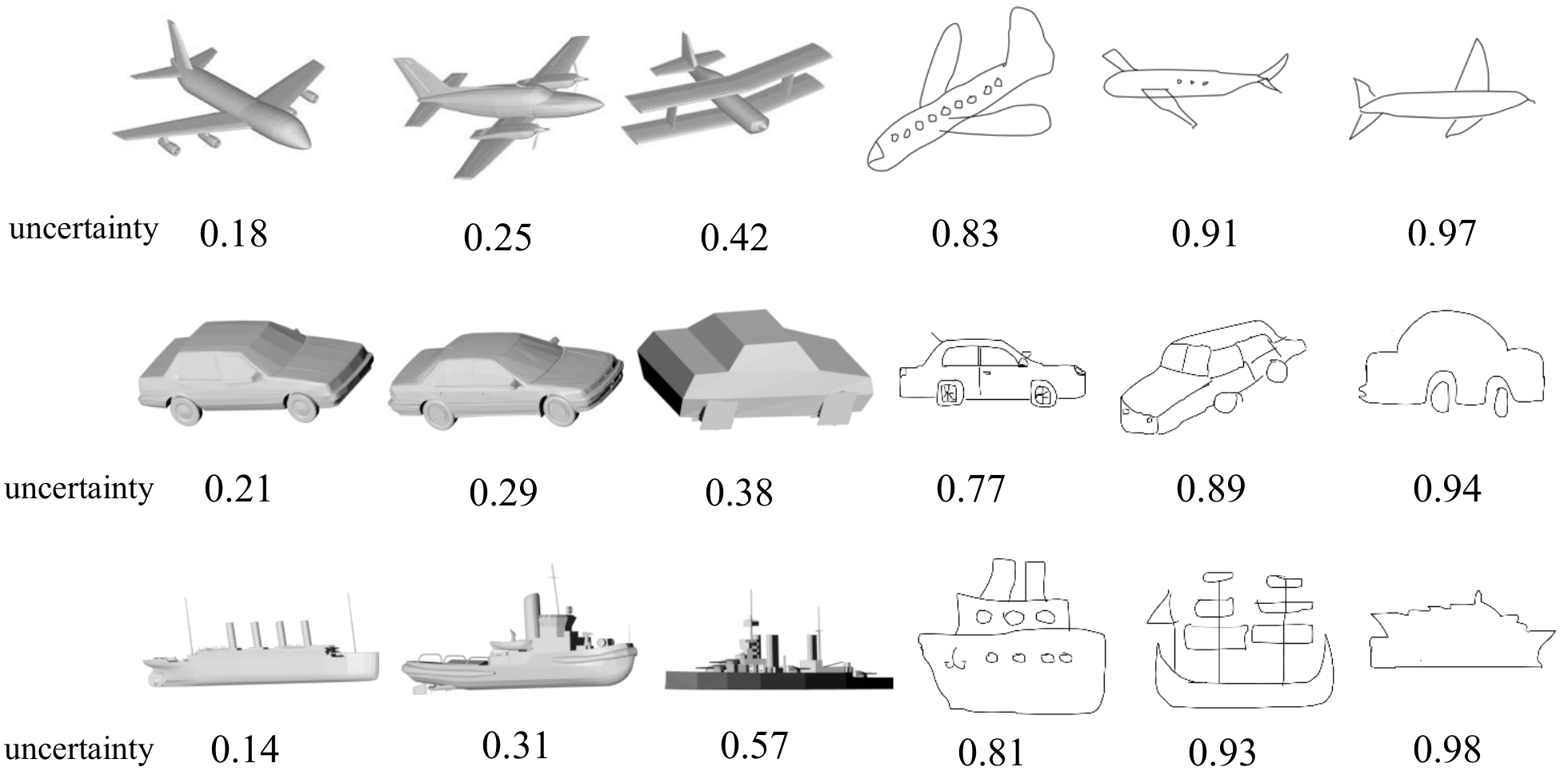}
\caption{The uncertainty predicted by CBUL with saimese network of sketches and 3D shapes examples from SHREC 2014. The model tends to predict smaller uncertainties for all 3D shapes and larger uncertainties for all sketches}
\label{fig3}
\end{figure}

\section{Experiments}

\subsection{Experimental Settings}

\textbf{Datasets.} We conduct experiments on two common benchmarks, SHREC 2013~\cite{2013SHREC} and 2014~\cite{2014SHREC}. SHREC 2013 contains 1258 3D shapes and 7200 hand-drawn sketches, grouped into 90 classes. Each class has 80 sketches, with 50 for training and 30 for testing. SHREC 2014 has a similar structure but is larger in scale, with 171 classes, 8987 3D shapes, and 13,680 sketches. Each class has 80 sketches, with 50 for training and 30 for testing. Due to more semantically similar categories and larger intra-class variations, SHREC 2014 is more challenging than SHREC 2013.

\noindent \textbf{Evaluation metrics.} Six common metrics~\cite{li2014comparison} are used for the evaluation of SBSR, including nearest neighbor (NN), first tier (FT), second tier (ST), E-measure (E), discounted cumulated gain (DCG) and mean average precision (mAP).

\noindent \textbf{Implementation details.} All of our experiments are implemented in PyTorch and run on an Nvidia RTX3090 GPU. For a fair and comprehensive comparison, we use ResNet50~\cite{he2016deep} and Inception-ResNet-v2 (I-R-V2)~\cite{szegedy2017inception}, both pretrained on ImageNet, as the backbones. The dimension of both sketch and shape embeddings for retrieval is 512. In pre-processing, all images are resized to $224 \times 224$ (ResNet50) / $299 \times 299$ (I-R-V2). We also use trivial augment~\cite{muller2021trivialaugment}, a type of automatic data augmentation, during training. The SGD optimizer is used with a batch size of 64. The initial learning rate is set to 4e-4, with a cosine annealing scheduler. The maximum number of training epochs is set to 200. These settings are the same for both sketch and 3D shape representation learning.

\begin{table}
  \caption{The performance (\%) on SHREC 2014. For each metric, the best result under the same backbone is in bold.}
  \label{tab2}
  \resizebox{0.48\textwidth}{!}{
  \begin{tabular}{llllllll}
    \hline
    Method&Backbone&NN&FT&ST&E&DCG&mAP\\
    \hline
    DPSML\cite{Lei2019-ki}&ResNet50&77.4&79.8&84.9&41.5&87.7&81.3\\
    \hline
    CGN\cite{dai2020cross}&ResNet50&78.9&81.1&85.0&41.8&88.1&83.0\\
    \hline
    JFLN\cite{zhao2022jfln}&ResNet50&79.2&82.3&84.7&42.4&87.3&83.3\\
    \hline
    \multirow{2}{*}{DSSH\cite{2019hashing} } &ResNet50&77.5&78.8&83.1&40.4&87.0&80.6\\
    &I-R-v2&79.6&81.3&85.1&41.2&88.1&82.6\\
    \hline
     \multirow{2}{*}{HEAR\cite{chen2020learning}} &ResNet50&79.2&80.7&84.6&40.9&87.8&82.2\\
    &I-R-v2&80.9&82.6&86.3&41.4&89.0&83.4\\
    \hline
    \multirow{2}{*}{SUL\cite{Liang2021-kg}} 
    &ResNet50&79.4&81.9&86.3&41.8&88.9&83.4\\
    &I-R-v2&81.1&82.9&87.1&42.0&89.5&83.9\\
    \hline
    \multirow{2}{*}{\makecell[l]{UACTN\\(Ours)}}
    &ResNet50&\textbf{81.0}&\textbf{83.7}&\textbf{86.9}&\textbf{42.7}&\textbf{89.2}&\textbf{84.8}\\
    &I-R-v2&\textbf{82.3}&\textbf{84.6}&\textbf{88.1}&\textbf{43.1}&\textbf{90.2}&\textbf{85.5}\\
    \hline
\end{tabular}}
\end{table}

\begin{figure}[htbp]
\includegraphics[width=0.48\textwidth]{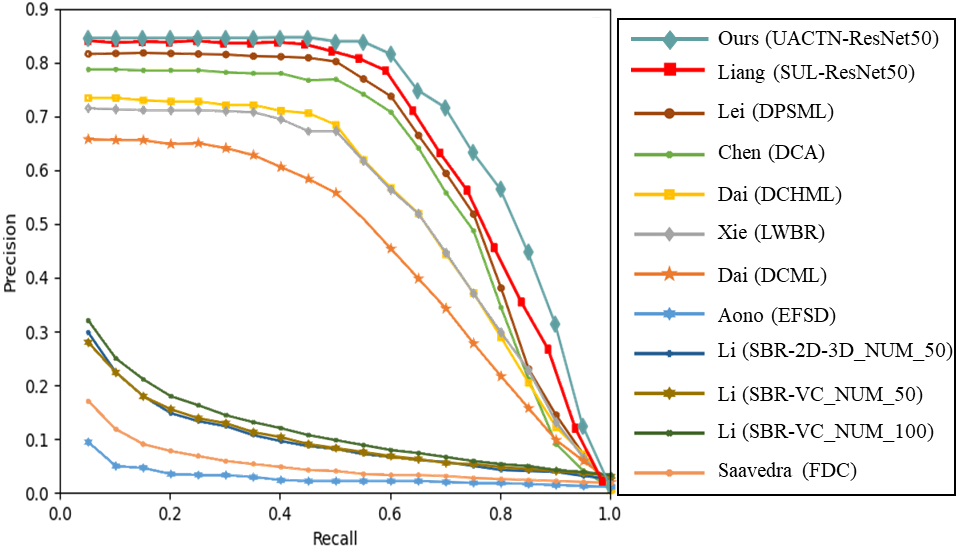}
\caption{Precision-recall curves of various method on SHREC 2013.}
\label{fig13}
\end{figure}

\begin{figure}[htbp]
\includegraphics[width=0.48\textwidth]{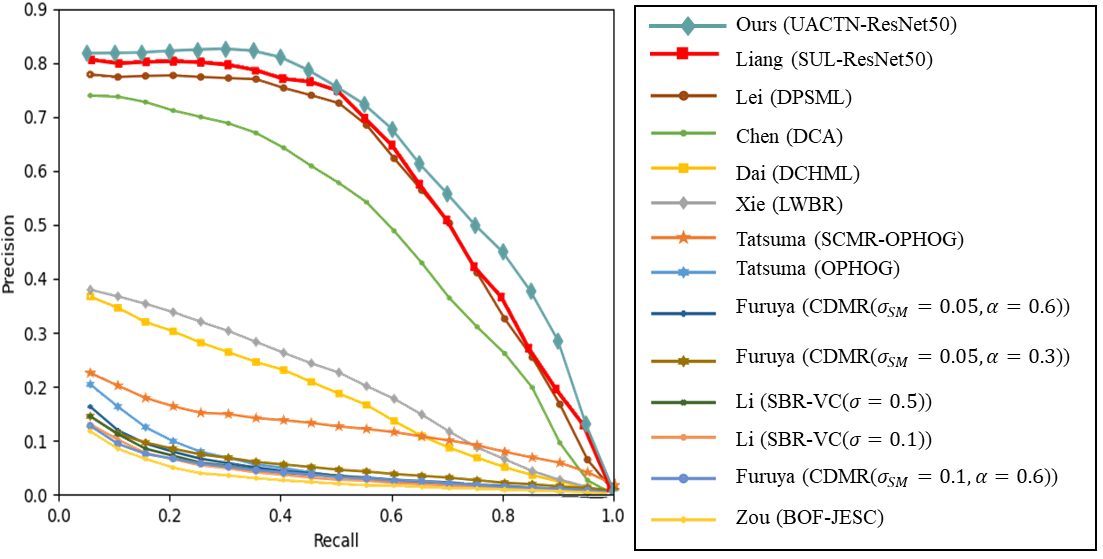}
\caption{Precision-recall curves of various method on SHREC 2014.}
\label{fig14}
\end{figure}

\begin{table}
  \caption{Ablation study on SHREC 2014 with ResNet50.}
  \centering
  \label{tab3}
  \resizebox{0.42\textwidth}{!}{
  \begin{tabular}{ccl}
    \hline
    Cross-modal matching&Uncerainty learning&mAP\\
    \hline
    \textit{siamese}~\cite{Liang2021-kg}&-&82.6\\
    \textit{siamese}~\cite{Liang2021-kg}&SUL~\cite{Liang2021-kg}&83.4\\
    \textit{siamese}~\cite{Liang2021-kg}&CBUL (Ours)&83.5\\
    \textit{transfer} (Ours)&-&83.6\\
    \textit{transfer} (Ours)&SUL~\cite{Liang2021-kg}&84.3\\
    \textit{transfer} (Ours)&CBUL (Ours)&\textbf{84.8}\\
    \hline
\end{tabular}}
\end{table}

\subsection{Comparison with the State-of-the-Art}

Tables~\ref{tab1} and~\ref{tab2} compare our UACTN with several state-of-the-art methods on the SHREC 2013 and 2014 datasets. And the precision-recall curves on the two datasets compared with several methods~\cite{Liang2021-kg,Lei2019-ki,Chen2018-up,Dai2018-ce,dai2017deep,2013SHREC,2014SHREC,li2015comparison,furuya2013ranking} are presented in Fig.~\ref{fig13} and Fig.~\ref{fig14}. It can be seen that the proposed UACTN outperforms these state-of-the-art methods for almost all evaluation metrics under the same CNN backbones on both datasets. For example, our method outperforms the current best method JFLN~\cite{zhao2022jfln} by 0.7\% mAP with ResNet50 and beats SUL~\cite{Liang2021-kg} by 1.5\% mAP with I-R-V2 on SHREC 2013. Our method also outperforms SUL~\cite{Liang2021-kg} with 1.4\% mAP with ResNet50 and 1.6\% mAP with I-R-V2 on SHREC 2014. Even compared with the best I-R-V2 backbone results achieved by SUL~\cite{Liang2021-kg}, our ResNet50 results exceed them by 0.2\% mAP on SHREC 2013 and 0.9\% mAP on SHREC 2014. These results demonstrate the superiority of our method, and the more significant advantage on SHREC 2014 shows that our approach is more effective in datasets with more classes and noisy samples.

\begin{figure}[tbp]
\centering
\includegraphics[width=0.48\textwidth]{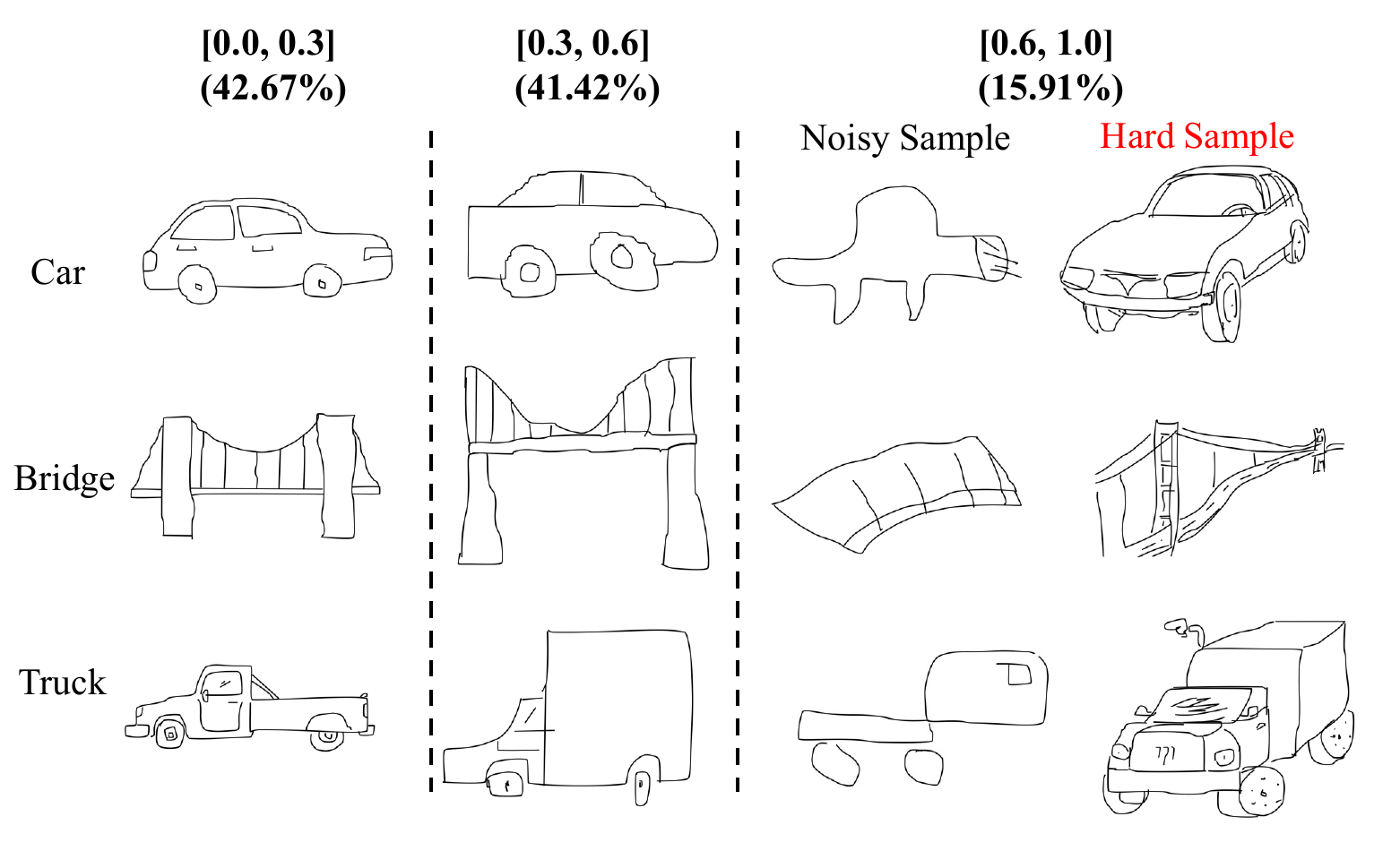}
\caption{Sketch examples from SHREC 2013 in three uncertainty intervals. The percentage of each interval is also shown.}
\label{fig4}
\end{figure}

\subsection{Ablation Study}

\noindent\textbf{Effect of the proposed modules.} Our contribution involves an end-to-end classification-based sketch uncertainty learning approach and a two-stage cross-modal matching framework based on 3D shape feature transfer, which are denoted by CBUL and \textit{transfer} respectively. Correspondingly, the traditional one-stage cross-modal matching framework based on the siamese network, which is the baseline in SUL~\cite{Liang2021-kg}, is denoted by \textit{siamese}. Moreover, for a fair comparison with SUL~\cite{Liang2021-kg}, which is also based on uncertainty learning, we re-implement SUL on \textit{transfer} and \textit{siamese}. Table~\ref{tab3} shows that each of the two proposed modules improves the retrieval performance. More detailed experimental results and network architecture of the methods in Table~\ref{tab3} are put in the Appendix.

The results suggest two observations. First, the proposed \textit{transfer} method improves the mAP by 1.0\% over the common \textit{siamese} method, even without uncertainty learning. This is because \textit{transfer} does not use shared network layers for sketch and 3D shape representation models, allowing for the representation learning of both modalities to be independent of each other. Second, the proposed CBUL does not demonstrate a clear advantage over SUL on \textit{siamese}, but CBUL outperforms SUL by 0.5\% mAP on \textit{transfer}. This is because the sketch and 3D shape representations are learned together on \textit{siamese} + CBUL, making it difficult to accurately represent sketch noise levels with the uncertainty as both sketch noise levels and modality gap between sketches and 3D shapes influence uncertainty learning. As illustrated in~\ref{fig3}, the model will tend to predict small uncertainties for 3D shapes and large uncertainty for sketches, which limits the ability of uncertainty to represent sketch quality. In contrast, SUL fine-tunes the trained sketch model and the 3D shape model is not involved in uncertainty learning on \textit{siames} + SUL, so the uncertainty learning is not disturbed by the modality gap. On \textit{transfer} + SUL and \textit{transfer} + CBUL, uncertainty learning in both cases is not affected by the modality gap, allowing CBUL to demonstrate its advantage of being trained from scratch.

\begin{figure}[tbp]
\includegraphics[width=0.48\textwidth]{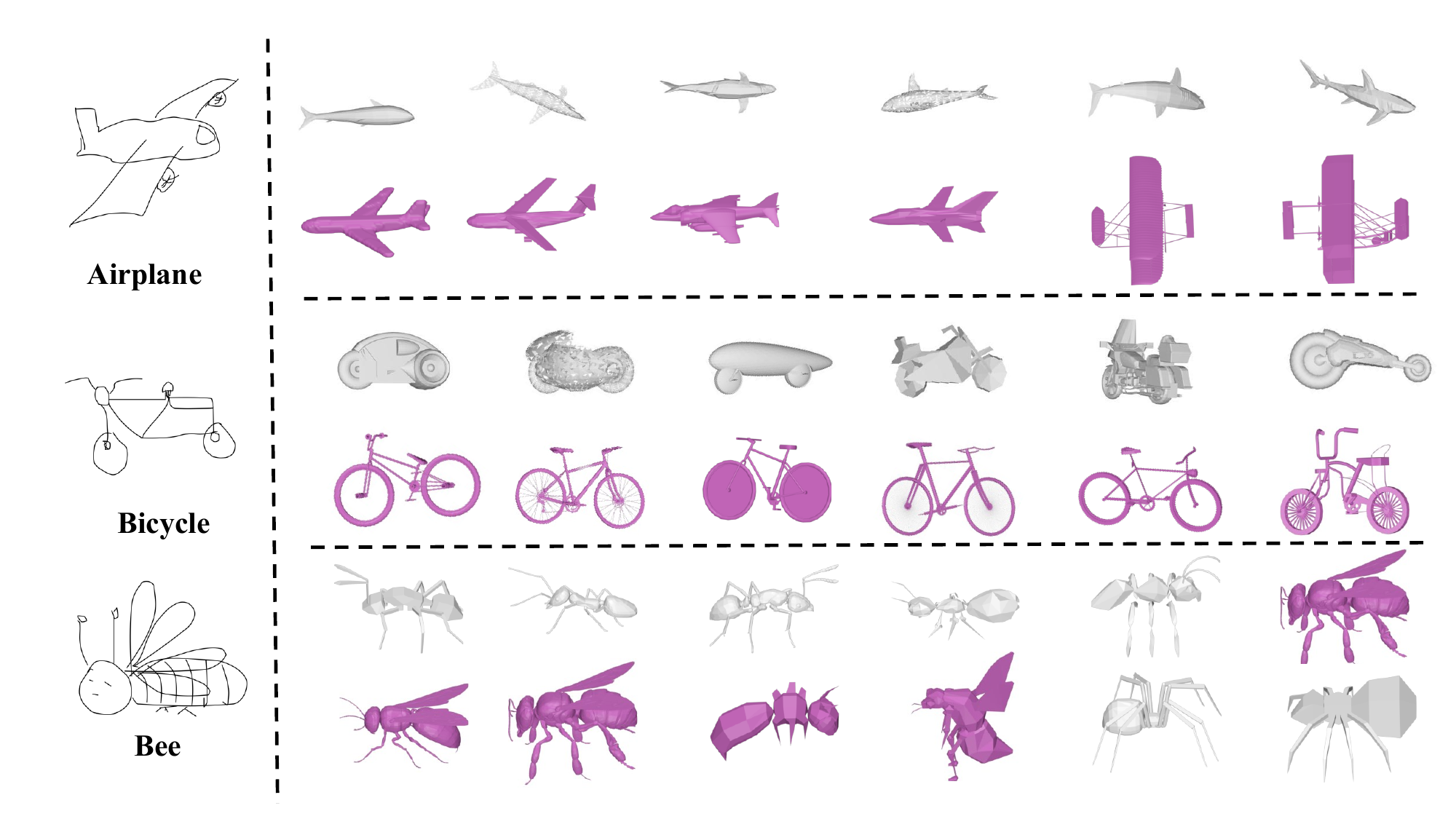}
\caption{Retrieval examples on SHREC 2014. For each sketch query, top row is the results of \textit{siamese} and bottom row corresponds to UACTN (\textit{transfer} + CBUL). Purple denotes the right retrieval results.}
\label{fig5}
\end{figure}

\noindent\textbf{Visualisation Results.} To verify the hypothesis that uncertainty can represent sketch noise levels, we visualize the estimated $\sigma^{2}_{i}$ for some examples from SHREC 2014. As $\sigma^{2}_{i}$ is a high-dimensional vector, we use the harmonic mean of each dimension as a measure of uncertainty. All uncertainty values are normalized to $(0,1)$ and separated into three intervals. As shown in Fig.~\ref{fig4}, sketches with lower uncertainty are typically easier to recognize, while most sketches with high uncertainty are lacking in detail and even unrecognizable. This suggests that the learned uncertainty values reflect the sketches' noise level. However, there are also counter-examples. Sketches rich in detail and easily recognizable to humans may differ significantly from other sketches in the dataset due to different drawing views. These samples may be assigned high uncertainty values even though they are not noisy. This is a limitation of our current approach, and we will develop an effective way to deal with unrecognizable noisy and recognizable hard samples separately in further research.

Fig.~\ref{fig5} shows some examples of retrieval results for the SHREC 2014 dataset using the \textit{siamese} and UACTN (\textit{transfer} + CBUL) methods. The results demonstrate that the proposed UACTN method achieves more promising results for the example classes compared to the \textit{siamese} method.

\section{Conclusion}

This paper proposes a novel Uncertainty-aware Cross-modal Transfer Network (UACTN) for sketch-based 3D shape retrieval. Our approach employs an end-to-end data uncertainty learning method on a two-stage cross-modal matching framework to prevent the model from overfitting to noisy sketches. In comparison to the work with a similar idea, we make more effective use of uncertainty information to improve sketch representation learning. Extensive experiments demonstrate the superiority of our method over state-of-the-art methods. In future work, we will investigate effective methods for dealing with recognizable hard samples.

\section*{Acknowledgment}

This work was supported in part by the National Natural Science Foundation of China under Grant 62076183, 61936014 and 61976159, in part by the Natural Science Foundation of Shanghai under Grant 20ZR1473500, in part by the Shanghai Science and Technology Innovation Action Project of under Grant 20511100700 and 22511105300, in part by the Shanghai Municipal Science and Technology Major Project under Grant 2021SHZDZX0100, and in part by the Fundamental Research Funds for the Central Universities. The authors would also like to thank the anonymous reviewers for their careful work and valuable suggestions.

% \begin{thebibliography}{00}
% \bibitem{b1} G. Eason, B. Noble, and I. N. Sneddon, ``On certain integrals of Lipschitz-Hankel type involving products of Bessel functions,'' Phil. Trans. Roy. Soc. London, vol. A247, pp. 529--551, April 1955.
% \bibitem{b2} J. Clerk Maxwell, A Treatise on Electricity and Magnetism, 3rd ed., vol. 2. Oxford: Clarendon, 1892, pp.68--73.
% \bibitem{b3} I. S. Jacobs and C. P. Bean, ``Fine particles, thin films and exchange anisotropy,'' in Magnetism, vol. III, G. T. Rado and H. Suhl, Eds. New York: Academic, 1963, pp. 271--350.
% \bibitem{b4} K. Elissa, ``Title of paper if known,'' unpublished.
% \bibitem{b5} R. Nicole, ``Title of paper with only first word capitalized,'' J. Name Stand. Abbrev., in press.
% \bibitem{b6} Y. Yorozu, M. Hirano, K. Oka, and Y. Tagawa, ``Electron spectroscopy studies on magneto-optical media and plastic substrate interface,'' IEEE Transl. J. Magn. Japan, vol. 2, pp. 740--741, August 1987 [Digests 9th Annual Conf. Magnetics Japan, p. 301, 1982].
% \bibitem{b7} M. Young, The Technical Writer's Handbook. Mill Valley, CA: University Science, 1989.
% \end{thebibliography}

\bibliographystyle{IEEEtran}
\bibliography{ref}

% \vspace{12pt}
% \color{red}
% IEEE conference templates contain guidance text for composing and formatting conference papers. Please ensure that all template text is removed from your conference paper prior to submission to the conference. Failure to remove the template text from your paper may result in your paper not being published.

\end{document}